\documentclass[a4paper]{spie}  
\usepackage[]{graphicx}
\usepackage[table]{xcolor}
\RequirePackage{amsmath}
\RequirePackage{xspace}
\RequirePackage{bbm}

\def\XS{\xspace}
\DeclareMathAlphabet{\mathb}{OML}{cmm}{b}{it}
\def\sbm#1{\ensuremath{\mathb{#1}}}                
\def\sbmm#1{\ensuremath{\boldsymbol{#1}}}          
\def\sdm#1{\ensuremath{\mathrm{#1}}}               
\def\scu#1{\ensuremath{\mathcal{#1\XS}}}           
\def\sbl#1{\ensuremath{\mathbbm{#1}}}              
\def\thetab      {{\sbmm{\theta}}\XS}      
\def\Sc{{\scu{S}}\XS}   
\def\ie{\textit{i.e.,}\XS}
\def\QD{{\sdm{Q}}\XS}  
\def\DD{{\sdm{D}}\XS}  

\def\beq#1\eeq{\begin{equation}#1\end{equation}}
\def\cro#1{\left[#1\right]}                
\def\btabu{\begin{tabular}}             \def\etabu{\end{tabular}}
\def\eeq{\end{equation}}
\def\Rbb{{\sbl{R}}\XS}  
\def\eR{\Rbb}
  
\def\Ib{{\sbm{I}}\XS}  
\def\eeq{\end{equation}}
\def\argmax{\mathop{\mathrm{arg\,max}}} 

\title{Comparative study of image registration techniques for bladder video-endoscopy} 

\author{Achraf {Ben-Hamadou}, Charles Soussen, Walter Blondel, Christian Daul and Didier Wolf
\skiplinehalf
Centre de Recherche en Automatique de Nancy (CRAN, UMR 7039, Nancy-University, CNRS),2, avenue de la for\^et de Haye, F-54516 Vand\oe uvre-l\`es-Nancy.}

\authorinfo{E-mail: \texttt{Achraf.Ben-Hamadou@ensem.inpl-nancy.fr,
    Charles.Soussen@cran.uhp-nancy.fr,
    Christian.Daul@ensem.inpl-nancy.fr,
    Walter.Blondel@ensem.inpl-nancy.fr,
    Didier.Wolf@ensem.inpl-nancy.fr}.}

\begin{document} 
  \maketitle 

\begin{abstract}
Bladder cancer is widely spread in the world. Many adequate diagnosis techniques exist. Video-endoscopy remains the standard clinical procedure for visual exploration of the bladder internal surface. However, video-endoscopy presents the limit that the imaged area for each image is about nearly $1~cm^2$. And, lesions are, typically, spread over several images. The aim of this contribution is to assess the performance of two mosaicing algorithms leading to the construction of panoramic maps (one unique image) of bladder walls. The quantitative comparison study is performed on a set of real endoscopic exam data and on simulated data relative to bladder phantom.
\end{abstract}

Robustness and accuracy


\section{Introduction}
The applicative aim of this contribution concerns bladder cancer detection in image
sequences recorded during endoscopic examinations. The 2-D cartography of an image sequence, also called image mosaicing, relies on a prior registration of consecutive image pairs of the video sequence, and then on the superposition of all the images onto a single common panoramic image. Lesion detection and evolution
assessment may be far easier in such mosaics than in isolated images showing, each, only a very small part of the region of interest. Mosaicing of human organ images is a few treated problem (see~\cite{Jalink:1996,Chou:1997,Can:2002a,Vercauteren:2005} for applications of mosaicing in mammography, angiography, ophthalmology, and microscopy), the existing solutions being not automated or needing {\em a priori} knowledge like sensor position and being only able to register few images. In the case of bladder endoscopy, image mosaicing is difficult for several reasons. First, image primitives are not easy to extract robustly (e.g., contours), and their background is severely textured. Moreover, the recorded images have a great inter- and intra-patient variability. Second, the endoscope position is unknown during the image acquisition, since urologists can move ``freely" the instrument inside the bladder. Third, a video sequence consists generally of thousands of images. One of the technical question for the consecutive registration of pairs of images is : how to register robustly,  precisely and with an acceptable computation time all the images of a sequence? The computation time may be the less critical factor since the mosaic must be available for a further diagnosis which is usually performed some dozens of minutes or hours after the examination itself.
In this paper, we focus on the registration of consecutive images, denoted by $\Ib_k$ 
(target image) and $\Ib_{k+1}$ (source image), where $k$ stands for the image index in the video sequence. The registration of $\Ib_k$ and $\Ib_{k+1}$ consists in finding a 2-D/2-D perspective transformation $T(x,y;\thetab_k)$ which superimposes $\Ib_{k+1}$ on $\Ib_k$. In notation 
$T(x,y;\thetab_k)$, $(x,y)$ represents a 2-D point in the domain of image $\Ib_{k+1}$ and $\thetab_k$ is the set of $a_{ij}$ parameters of the perspective transformation (related in eq.~\eqref{eq:trans} to the translations $t_x$ and $t_y$, plane rotation $\phi$, scale factor $f$, shearing parameters $S_x$ and $S_y$, and perspective parameters $a_{31}$ and $a_{32}$). The perspective transformation $(x',y')=T(x,y;\thetab)$ of the 2-D space reads :
\beq
\cro{\btabu{c}$x'$\\$y'$\etabu}\;=\;\frac{1}{w}\:
\cro{\btabu{c}$u$\\$v$\etabu},\;\;\;\textrm{where}\;\;\;\;
\cro{\btabu{c}$u$\\$v$\\$w$\etabu}\;=\;
\cro{\btabu{ccc}
$\underbrace{f\cos(\phi)}_{a_{11}}$&$\underbrace{-S_x\sin(\phi)}_{a_{12}}$&$\underbrace{t_x}_{a_{13}}$\\
$\underbrace{S_y\sin(\phi)}_{a_{21}}$&$\underbrace{f\cos(\phi)}_{a_{22}} $&$\underbrace{t_y}_{a_{23}}$\\
$a_{31}$&$a_{32}$&$a_{33}$ \etabu }\:
\cro{\btabu{c}$x$\\$y$\\$1$\etabu} \label{eq:trans} \eeq
and involves 8 independent parameters ($a_{33} = 1$). The registration of images $\Ib_k$ and $\Ib_{k+1}$ is stated as the maximization of a similarity criterion of the form:
\beq\label{eq:simil} 
{\thetab}_k\:=\;\argmax_{\thetab\in\eR^8}\,\Sc(\Ib_{k},\,T(\Ib_{k+1};\thetab)).
\eeq
The difference between different registration algorithms to be chosen to solve this problem lies in the
choice of the measure of similarity \Sc and in the choice of the numerical algorithm of optimization.

\section{Image registration algorithms}
The bladder images do not systematically include image primitives (e.g., corners or contours) that can be robustly enough extracted\cite{Yahir:2007}. For this reason, the most simple registration methods relying on the segmentation of an image primitive cannot be used, and we must consider a great number of image pixels when choosing the measure of similarity $\Sc(\Ib_{k},\,T(\Ib_{k+1};\thetab))$.

\subsection{$A_{QD}$ : Quadratic distance based algorithm}
The first algorithm $A_{QD}$\cite{Yahir:2006,Yahir:2007} is based on a measure of dissimilarity $\Sc_{\QD\DD}(\Ib_{k},\,T(\Ib_{k+1};\thetab))$ defined as the quadratic distance between the grey levels of the pixels of $\Ib_{k}$ and these of the perspective transformation of the pixels of $\Ib_{k+1}$ :

\begin{equation}
\Sc_{\QD\DD}(\Ib_{k},\,T(\Ib_{k+1};\thetab)) = \sum_{(x,y)\in \Ib_{k}\cap\Ib_{k+1}}\left[\Ib_{k}(x,y) - \Ib_{k+1}(T(x,y;\thetab)) \right]^2 
\end{equation}

where ($x,y$) denotes the coordinates of a pixel common to both $\Ib_{k}$ and $T(\Ib_{k+1};\thetab)$ images.
The minimization of this measure can be done using Baker and Matthews' inverse composition algorithm \cite{Baker:2004} whose goal is to estimate the optical flow, \ie the apparent motion between two given images.

\subsection{$A_{MI}$ : Mutual Information based algorithm}

The second algorithm $A_{MI}$\cite{Miranda:2008,Miranda:2005} is based on Viola and Wells' approach
 EMMA\cite{viola:1997} (EMpirical entropy Manipulation and Analysis). $A_{MI}$ aligns images $\Ib_{k}$ and $\Ib_{k+1}$ by maximizing the measure of similarity $\Sc_{MI}(\Ib_{k},\,T(\Ib_{k+1};\thetab))$ defined as the mutual information between $\Ib_{k}$ and $T(\Ib_{k+1};\thetab)$. Shortly speaking, the mutual information is a statistical measure computed with the grey level entropies $H(\Ib_{k})$ and $H(T(\Ib_{k+1};\thetab))$ of the overlapping parts of $\Ib_{k}$ and $T(\Ib_{k+1};\thetab)$ and with the joint entropy $H(\Ib_{k},T(\Ib_{k+1};\thetab))$ : 
\begin{equation}
\Sc_{MI}(\Ib_{k},\,T(\Ib_{k+1};\thetab)) = H(\Ib_{k}) + H(T(\Ib_{k+1};\thetab)) - H(\Ib_{k},T(\Ib_{k+1};\thetab))
\label{eq:mi}
\end{equation}

This measure is used together with a stochastic descent gradient algorithm in the optimization process of eq.~\eqref{eq:simil}. The mutual information is well suited to the registration of textured images \cite{PluimMV:03}.

\section{Comparative study : experiments and results}
In this section, we present the registration results obtained with both measures of similarity applied on common data sets obtained from real human bladder examinations and simulated endoscope displacement and on simulated data from a realistic phantom constructed using a pig bladder cartography.
\subsection{Robustness evaluation}
\begin{figure}[h]
   \begin{center}
   \begin{tabular}{ccc}
\includegraphics[height=4cm]{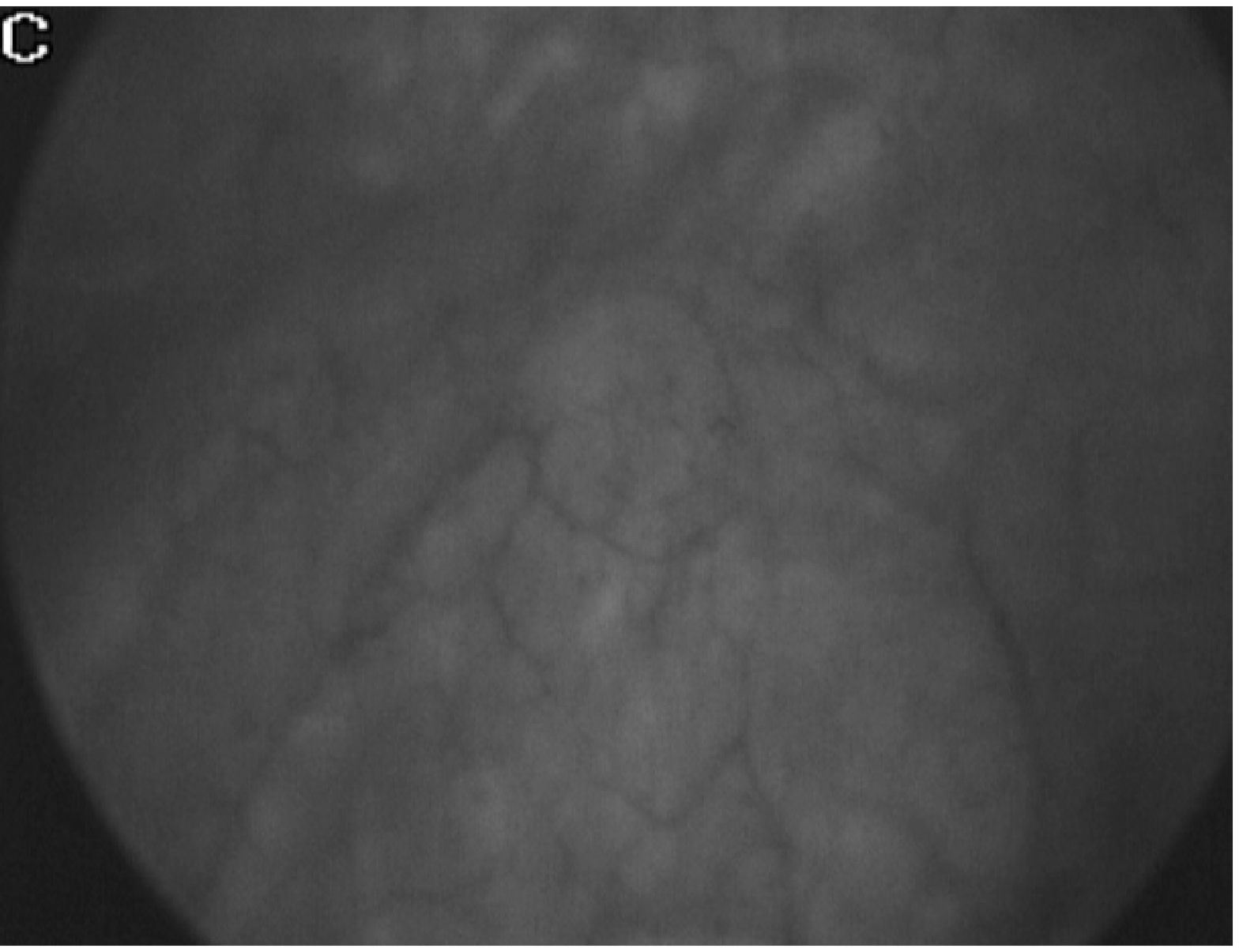} & \includegraphics[height=4cm]{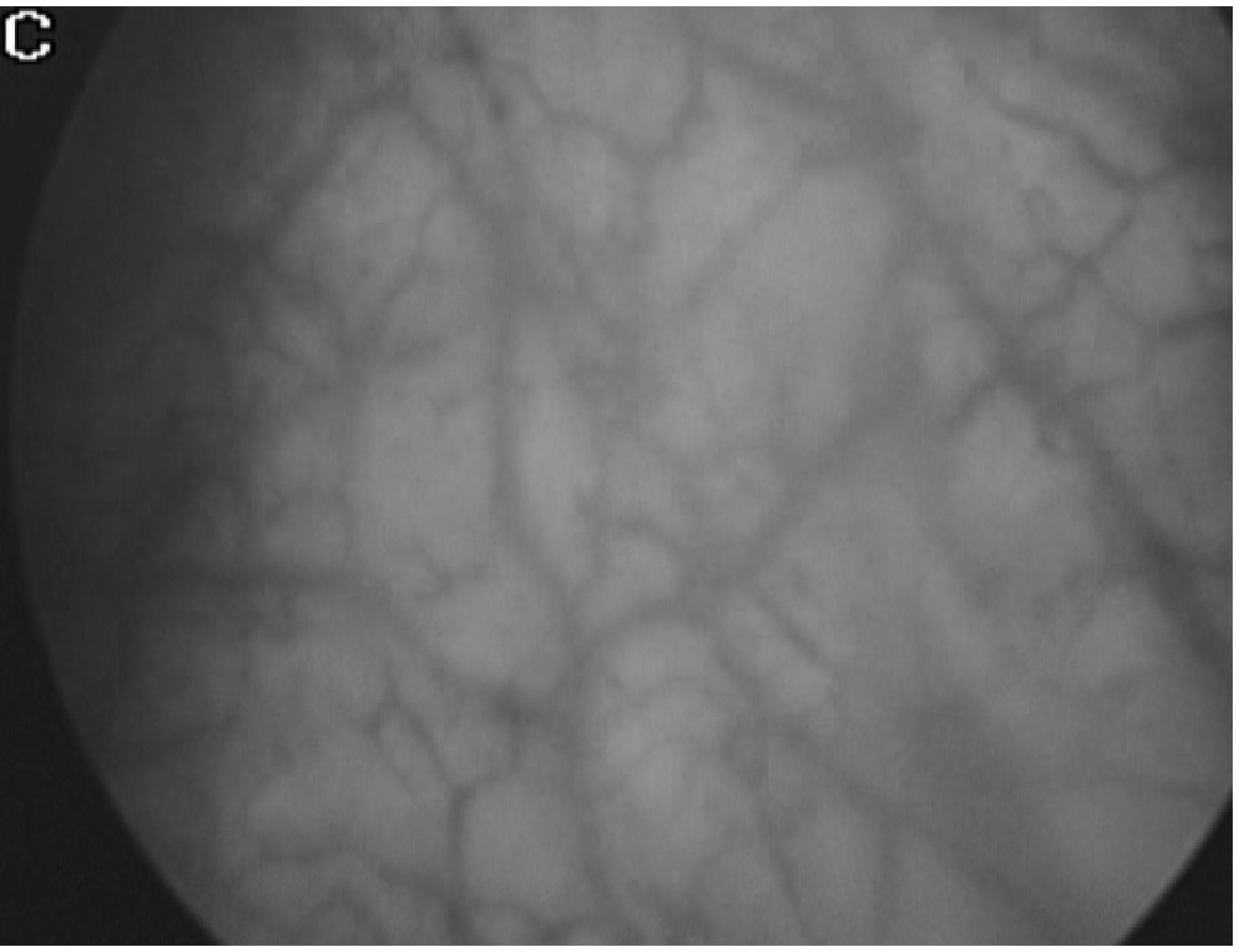} & \includegraphics[height=4cm]{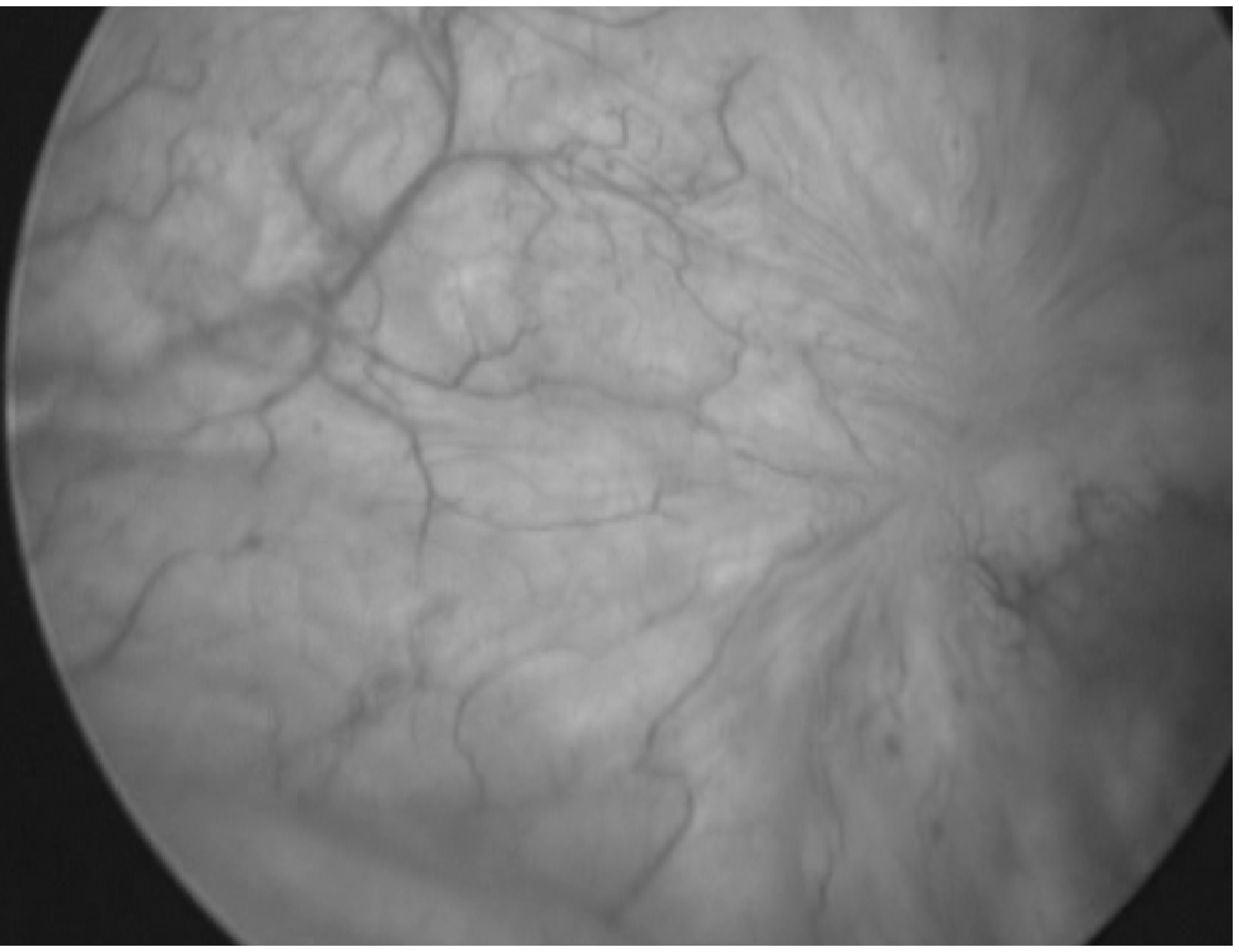} \\ 
 I & II & III\\
   \end{tabular}
   \end{center}
   \caption{I, II and III : Three reference images extracted from a real endoscopic exams for robustness evaluation tests. The chosen images present both texture and illumination variabilities.}
   \label{three Images} 
   \end{figure}

Three images with very different visual aspects (various textures and illumination conditions) were extracted from human endoscopic sequences to assess the robustness of the algorithms (see Figure~\ref{three Images}). These three images were all taken as reference images ($\Ib_k$ target images in eq.~\eqref{eq:simil}). $\Ib_{k+1}$ source images were computed by applying known simulated 2-D transformations on the $\Ib_k$ target image ($\Ib_{k+1} = T(\Ib_{k+1};\thetab)$), as if we simulate a real 3-D displacement of the endoscope. The 3-D displacement includes two translations corresponding to $t_x$ and $t_y$ in eq.~\eqref{eq:trans} while $t_z$ relates to the $f$ scale factor and 3-D rotations (in plane rotation $\phi$ in eq.~\eqref{eq:trans} and two out of plane rotations $\psi$ and $\alpha$ related to $a_{31}$ and $a_{32}$. In this way, it is possible to compare the calculated transformations with the known transformations already used to simulate images.

These ($\Ib_{k}$, $\Ib_{k+1}$) image pairs allow for an assessment of the largest endoscope viewpoint change leading to successful registrations. The parameter value intervals for which a successful registration was obtained are detailed in Tab.\ref{intervals}. For the $A_{MI}$ algorithm, intervals are : $t_x$ = $t_y$ = $\pm 30$ pixels, $f$ = $\pm 25\%$, $\phi$ = $\pm 20^{\circ}$ and $\alpha$ =$\psi$= $\pm 20^{\circ}$. These limits are more restricted for the $A_{QD}$ algorithm : $t_x$ = $t_y$ = $\pm 25$ pixels, $f$ = $\pm 15\%$, $\phi$ = $\pm 10^{\circ}$ and $\alpha$ = $\psi$= $\pm 10^{\circ}$. Even if for both methods, the translation limits are roughly the same order (with a slight advantage for the mutual information algorithm $A_{MI}$), {\em the robustness is clearly better for the mutual information method in terms of scale factor changes and in- and out of plane rotations}.\\
\begin{table}[h]
\begin{center}
\begin{tabular}{l|ccc}
  & Transformation value intervals &  &  \\ 
\hline Transformation parameters & $A_{QD}$ & $A_{MI}$ & Real endoscopic exam \\ 
\hline Translation ($t_x$ and $t_y$) & $\pm 25$ pixels & $\pm 30$ pixels &  $\pm 5$ pixels\\ 
 Scale factor ($f$) & $\pm 15\% $ & $\pm 25\%$ & $\pm 2\%$  \\ 
 In plane rotation ($ \phi$) & $\pm 10^{\circ}$ & $\pm 20^{\circ}$ & $\pm 1^{\circ}$ \\ 
 Out of plane rotations ($\psi$ and $\alpha$) & $\pm 10^{\circ}$ & $\pm 20^{\circ}$ & $\pm 1^{\circ}$\\ 

\end{tabular}
\end{center}
\caption{Transformation value intervals for with a successful registration was obtained for both $A_{QD}$ and $A_{MI}$. The last column designate transformation value intervals in real endoscopic exams in most cases ($90\%$).}
\label{intervals}
\end{table}   
\\
 
\subsection{Accuracy evaluation}
\begin{figure*}[h]
   \begin{center}
   \begin{tabular}{ccc}
\includegraphics[height=5cm]{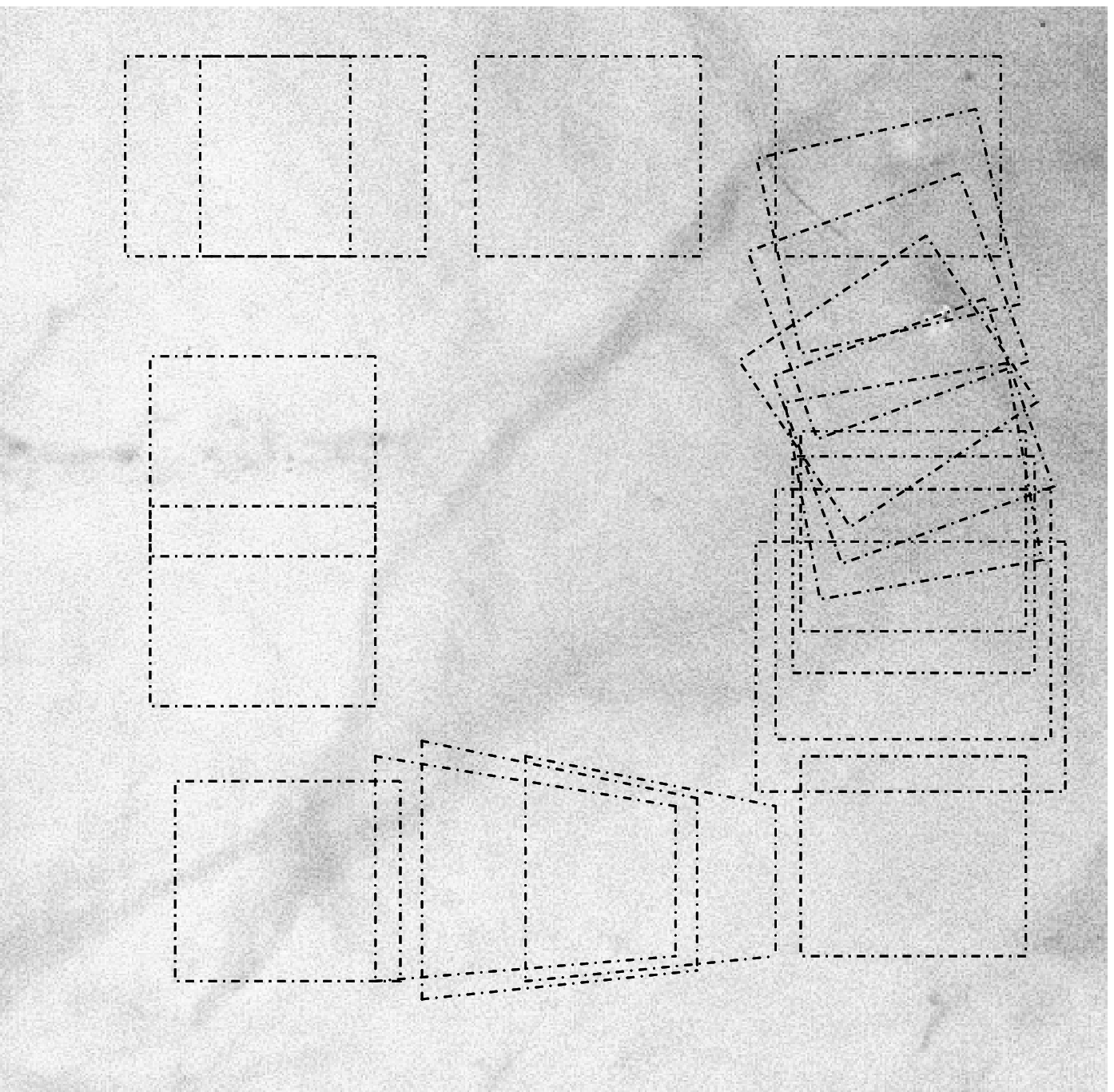} & \includegraphics[height=5cm]{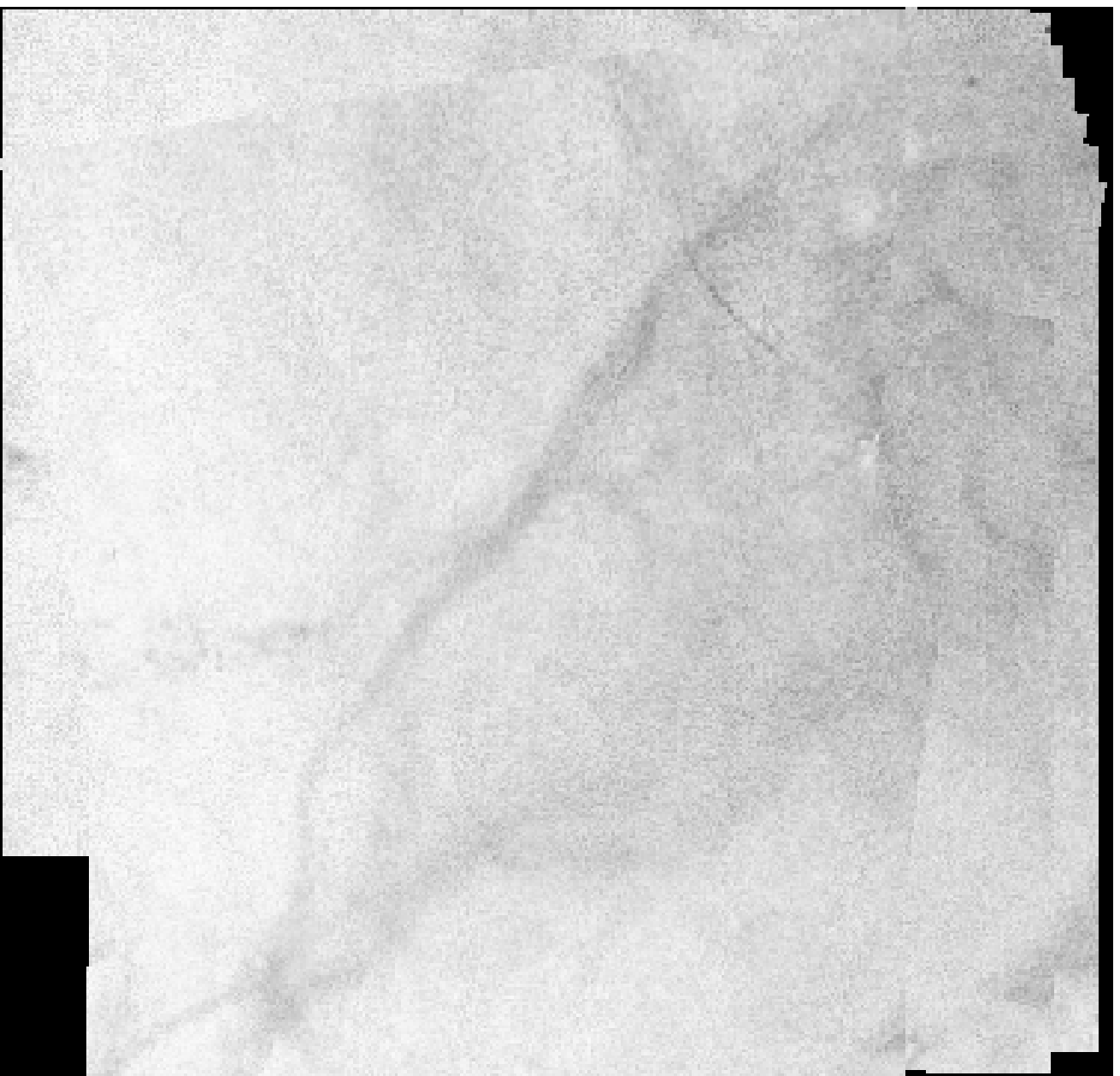} &  \includegraphics[height=5cm]{mosa_IM.eps} \\ 
(a) & (b) & (c)\\
 
   \end{tabular}
   \end{center}
   \caption{(a) Pig bladder photograph : the boxes indicate the simulated image sequence, \ie the acquisition path. (b) Mosaic (map) image obtained with the mutual information algorithm using registration of successive images. The map is visually coherent, all textures being continuous from one image to another in the map. This map visually matches the image of the pig bladder photograph. (c) Same results for the optical flow
method.}
   \label{results} 
   \end{figure*}

A quite realistic phantom was built using an excised pig bladder in order to test the registration accuracy of both methods. The pig bladder was incised, opened out and photographed with a camera. The pig bladder texture (see Figure~\ref{results}(a)) is very similar to that human bladder. The area covered by the acquired picture is a 16 cm side square. The first image was taken in the upper left photograph corner. The other images of the sequence were obtained by simulating successively 10 pixel horizontal  translations (14 upper images), a combination of 10 pixel translations and of  $2^{\circ}$ in plane rotations (upper 10 vertical images on the right photograph side), combination of 10 pixel translations and of 5\% scale factor changes (lower 10 vertical images on the right photograph side), a combination of 10 horizontal translations and of $4^{\circ}$ out of plane rotations (first 10 lower images from the right photograph side), {\em etc}.

All image pairs ($\Ib_{k}$, $\Ib_{k+1}$) were registered with both methods. The $\overline{\epsilon}_{k,k+1}$ registration accuracy criterion is defined as the mean distance between homologous pixels of the target images $\Ib_{k}$ and the registered images $T(\Ib_{k+1};\thetab_k)$. This criterion is ideally equal to 0. 

\begin{figure}[h]
   \begin{center}
	 \includegraphics{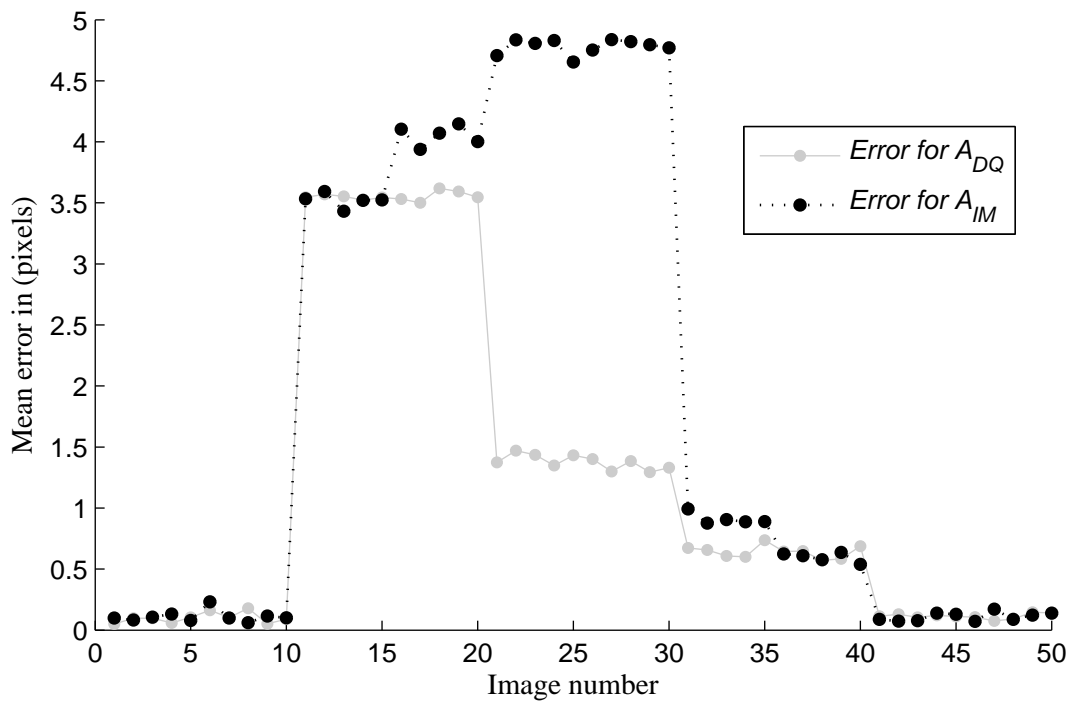}  
   \end{center}   
   \caption {Registration mean errors $\overline{\epsilon}_{k,k+1}$ for both $A_{QD}$ and $A_{MI}$. $\overline{\epsilon}_{k,k+1}$ values are equivalent for both algorithms in most sequence parts except in the part where the scale factor changes (images number 20 to 30) and for which $A_{QD}$ algorithm is more efficient.}
   \label{accuracy} 
   \end{figure}

In the case of simple translations ($\overline{\epsilon}_{k,k+1} \approx$ 0.2 pixels) and a combination of out of plane rotations (perspective changes) and translations ($\overline{\epsilon}_{k,k+1} \approx$ 0.6 pixels), the registration errors are equal for both methods. These errors are very small and imperceptible (see Figures~\ref{results}(b) and \ref{results}(c)). For the combinations of translations and in plane rotations, the errors are again equal for both methods ($\overline{\epsilon}_{k,k+1} \approx$ 3.5 pixels) (see Figure~\ref{accuracy}). 
As we observed visually (Figure \ref{results}), these errors rather correspond to a small $T(\Ib_{k+1};\thetab_k)$ image distortion without affecting the global mosaic (map) coherence. Especially in the map regions including image borders, the textures are without discontinuities. As shown in Figure \ref{accuracy}, registration mean errors $\overline{\epsilon}_{k,k+1}$ values are equivalent for both algorithms in most sequence parts except in the part where the scale factor changes (images number 20 to 30) and for which $A_{QD}$ algorithm is more efficient ($\overline{\epsilon}_{k,k+1} \approx$ 1.5 pixels compared to 4.5 pixels). Again, these errors do not affect the global visual map coherence. It is noticeable that, due to the image acquisition rate (25 images/second) and to the small endoscope displacements (few millimetres/second), the real rotation parameters ($<1^{\circ}$), translation parameters ($t_x$ and $t_y < $5 pixels) and scale factor ($<2\%$) changes are in fact by far smaller than those imposed in our experiments. In practice, both methods led systematically to sub-pixel errors for limited and more realistic displacements.

\begin{figure}[h]
   \begin{center}
	 \includegraphics[height=9cm]{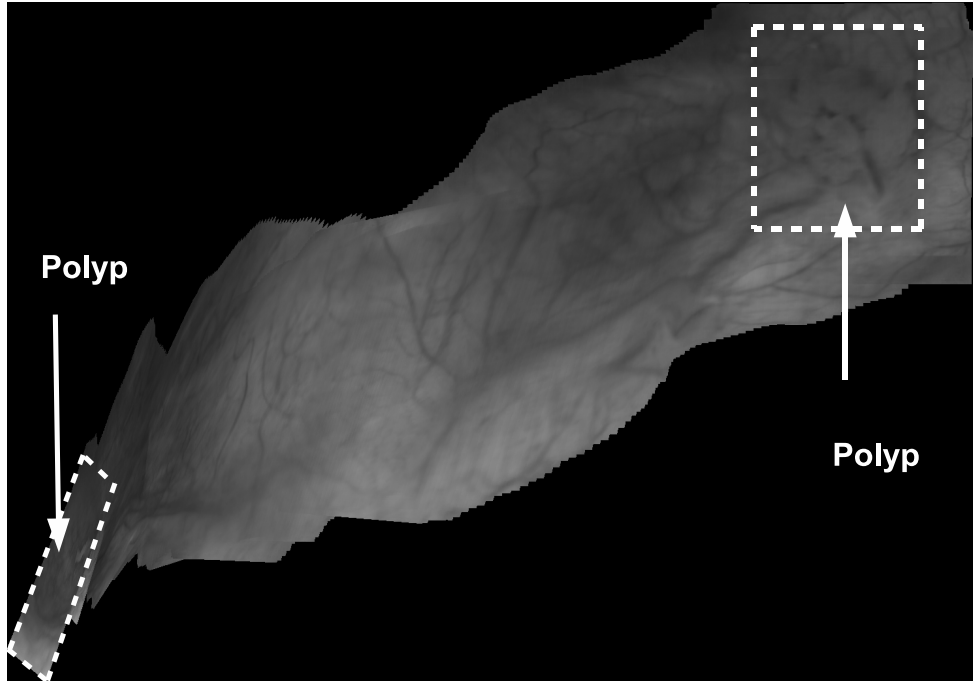}
   \end{center}   
   \caption{A 1479 $\times$ 1049 pixels panoramic image constructed from a 450 image sequence using $A_{QD}$. Two polyps are visible on the top-right and at the bottom left of the image. In this panoramic image, both polyps can be accurately located in relation to each other.}
   \label{pano_QD} 
   \end{figure}

\begin{figure}[h]
   \begin{center}
	 \includegraphics{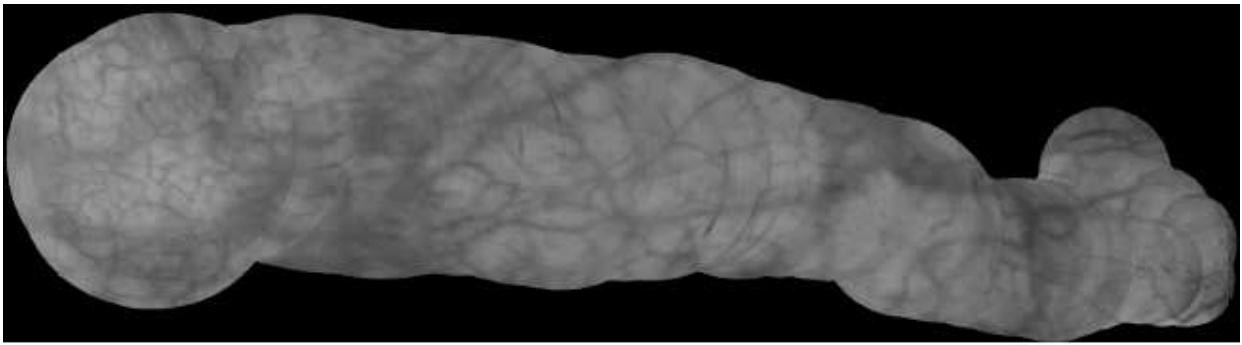}  
   \end{center}   
   \caption{A 650 $\times$ 182 pixels panoramic image constructed from a 500 image sequence using $A_{MI}$. In this panoramic image, there are no visible discontinuities on texture affirming a quite good visual coherence.}
   \label{pano_MI} 
   \end{figure}

Figures \ref{pano_QD} and \ref{pano_MI} show two panoramic images constructed from real cystoscopic examination images using $A_{QD}$ and  $A_{MI}$ respectively. The panoramic image in Figure \ref{pano_QD} is a 1479 $\times$ 1049 pixel image constructed from a 450 image sequence using $A_{QD}$. In this panoramic image, two polyps are visible on the top-right and at the bottom left of the image. Both polyps can be accurately located in relation to each other. Figure \ref{pano_MI} represents a 650 $\times$ 182 pixel panoramic image constructed from a 500 cystoscopic image sequence using $A_{MI}$. There are no visible discontinuities on texture affirming a quite good visual coherence.
%
%

\subsection{Mosaicing speed}

$A_{MI}$ and $A_{QD}$ were programmed in C language using OpenCV vision library. The evaluation of both algorithms robustness and accuracy was done using an Intel Dual core(TM) 2.40GHz, 2Gb RAM computer. The optimization method of the $A_{MI}$ algorithm requires, in average, 250 iterations to register consecutive images. And each image pair registration takes between 50 and 60 seconds. In Figure \ref{pano_MI}, the construction of the panoramic image took nearly 8 hours 27 minutes. However, in the same experimental conditions, $A_{QD}$ is about 100 times faster than $A_{MI}$. In fact, a mean number of 12 iterations was needed by the optimization algorithm of the $A_{QD}$ algorithm to register a pair of images. The time of registration for an image pair varied between 0.3 and 0.6 second. The panoramic image shown in Figure \ref{pano_QD} was constructed in 3.20 minutes. The computation time of the $A_{QD}$ makes possible the construction of partial panoramic image of the bladder during the standard cystoscopic examination procedure.
\section{Conclusion}
In terms of accuracy, both registration methods give comparable results with a slight advantage for $A_{QD}$. However, the $A_{MI}$ method is more robust than the $A_{QD}$ method, while the computation time of $A_{QD}$ algorithm (some tenth of seconds to register two images) is about 100 times smaller than that of the $A_{MI}$ algorithm. Future work will aim at combining both methods to reach the robustness of the mutual information method and to tend towards the computation times of the $A_{QD}$ algorithm. 


\acknowledgments
The authors express their gratitude to the ``R\'egion Lorraine" the ``Ligue Contre le Cancer (CD 52, 54)" and address their grateful thanks to physician Pr. F. Giollemin and urologist M.-A. DHallewin from Cancer Institute CAV in Nancy (France) for their clinical experience and for providing video sequences of various cystoscopic examination. The authors also thank the surgeons from Experimental Surgery Laboratory (Faculty of Medicine, Nancy) for the fresh pig bladders excisions.

\bibliography{biblio}

\end{document}